# Beyond Benchmarking: Scenario-Based Evaluation of Large Language Models for Personalized Learning

Bo Yuan[1] and Jiazi Hu[2]

[1] The University of Queensland, QLD 4072, Australia
[2] AI U Course, Kunming 650000, P.R. China
boyuan@ieee.org, jessica@aiucourse.com

**Abstract.** While large language models (LLMs) are increasingly being adopted to support personalized learning, there remains limited understanding of how their pedagogical behaviors differ in authentic learning scenarios. Existing evaluation practices often emphasize benchmark scores and overall model rankings, but such approaches usually provide limited insight into how LLMs diagnose student understanding and generate personalized guidance. This study proposes a scenario-based evaluation framework for closely examining LLM behavior in personalized learning support. Using a post-class tutoring setting as an illustrative example, a dataset comprising a student's responses to a set of data structures questions is provided to multiple LLMs. Each model is required to identify the underlying knowledge concepts, infer the student's mastery profile, and generate personalized guidance for improvement. To support consistent, reproducible and scalable comparison, Gemini is employed as an external evaluator across multiple pedagogically relevant dimensions, including diagnostic accuracy, instructional clarity, actionability, misconception identification, and appropriateness to the student's level. The resulting pairwise preferences are then fitted using the Bradley–Terry model to derive comparative strength estimates, while qualitative analysis and semantic visualization are used to further examine differences in feedback structure, diagnostic depth, and recommendation specificity. The key findings show that different LLMs exhibit distinguishable pedagogical behaviors within the same learning scenario and demonstrates how scenario-based evaluation can provide educationally meaningful signals for understanding model behavior in AI-enhanced education.

## 1 Introduction

Recently, large language models (LLMs) such as GPT-4o [1], DeepSeek-V3 [2], and GLM-4.5 [3] have demonstrated remarkable progress in natural language understanding and generation. Unlike traditional learning models, LLMs operate as generative models, capable of producing contextually coherent text, explanations, and even problem-solving strategies [4]. This transition from discriminative to generative paradigms marks a significant shift the AI landscape, enabling LLM-based systems not only to recognize patterns but also to generate multimodal outputs and engage in open-ended interaction. These advances have expanded the scope of AI applications and created new opportunities in education, where dynamic, adaptive, and personalized feedback has long been desired but difficult to achieve [5].

In traditional classroom settings, the time and cognitive resources of instructors are highly limited, making it challenging to provide individualized attention and tailor instruction, feedback, and pacing to individual learners' needs. Furthermore, assessment is often reduced to correctness scores, without deeper analysis of the underlying misconceptions that shape performance. As a result, feedback often remains generic and uniform, which limits students' opportunities to receive targeted support. This lack of personalization can reduce learning efficiency, hinder conceptual understanding, and diminish students' motivation and interest [6, 7].

Despite the growing enthusiasm, systematic evidence on how LLMs should be understood and evaluated in educational settings remains limited [8] and much of the current discourse emphasizes general benchmark performance such as factual knowledge, reasoning, mathematics, and coding [9]. While such benchmarks may provide useful indicators of general model capability, they offer limited insight into how LLMs behave when supporting authentic educational activities. From an educational perspective, a model that performs well on general-purpose benchmarks does not necessarily exhibit the same level of pedagogical competency when diagnosing student misconceptions, inferring mastery levels, or generating personalized learning guidance.

Existing studies in AI-enhanced education often focus on a single model within a specific context that is not sufficiently standardized or disclosed, making it difficult for other researchers to reproduce experiments or conduct fair comparisons [10, 11]. Such practices may hinder cumulative progress, as results remain fragmented across prompts, datasets, and instructional settings. Moreover, many existing evaluations rely primarily on student performance outcomes or expert judgments [12, 13]. While informative, these measures are inherently limited: student performance is potentially influenced by many uncontrolled factors beyond the tutoring intervention, and expert judgments are susceptible to subjectivity and bias [14].

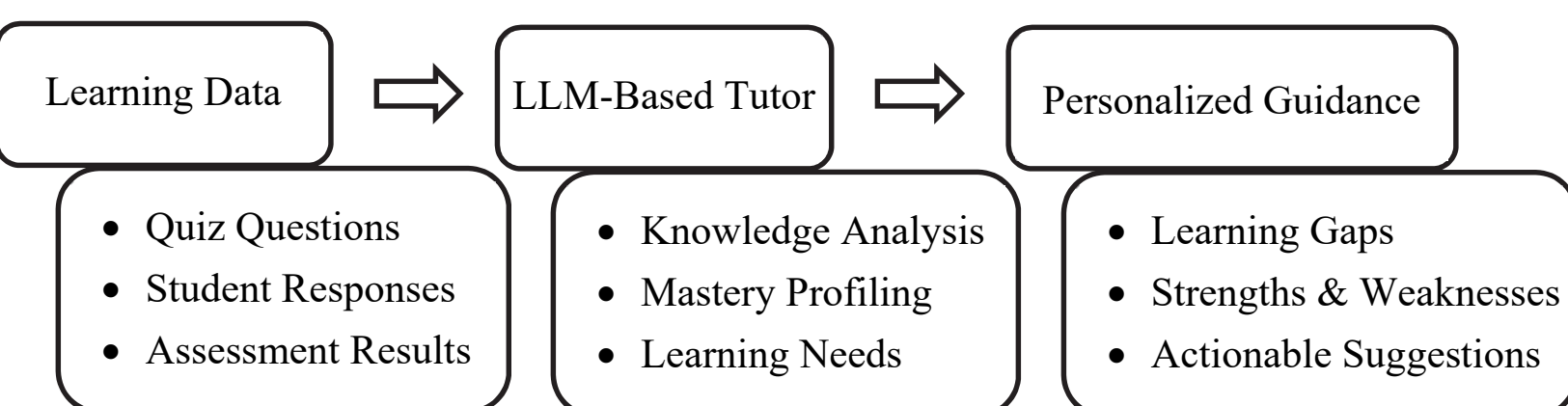

**Fig. 1.** Illustration of personalized learning with LLMs, showing how LLMs can transform student learning evidence into knowledge diagnosis, mastery inference, and personalized guidance.

Rather than asking which model is universally superior, this study investigates how different LLMs behave when placed in the same personalized learning scenario and how such differences can be systematically evaluated. Specifically, we propose a scenario-based evaluation framework grounded in a realistic tutoring setting. The learning data of students, including quiz items and responses, along with correctness labels, are provided to multiple LLMs, each of which is required to identify underlying knowledge concepts, infer mastery levels, and generate personalized learning guidance (Fig. 1). To ensure comparability, all models operate under a standardized prompting protocol, thereby minimizing the variations introduced by prompt engineering. Meanwhile, Gemini [15] is employed as a consistent external evaluator [16, 17] to conduct pairwise assessment across multiple pedagogically relevant dimensions.

Experimental results illustrate that different LLMs exhibit distinct pedagogical behaviors in personalized learning support. More importantly, the study demonstrates how scenario-based evaluation workflows can generate more educationally meaningful assessment signals than benchmark-oriented evaluation alone. In this sense, the proposed framework provides a transparent and extensible methodology for future empirical research on LLM evaluation in AI-enhanced education.

## 2 Background

Personalized learning refers to an instructional approach in which teaching strategies, learning resources, and pacing are adapted to the individual needs, abilities, and interests of each student. The value of personalized learning has been widely recognized in both educational theory and practice [18]. At its core, personalization enhances learning effectiveness by aligning instruction with students' existing knowledge and skills rather than constraining them to the average pace of a classroom. It also supports the growth of self-regulated and lifelong learning skills, as students are encouraged to reflect on their progress and take ownership of their learning trajectories.

Historically, individualized tutoring was reserved for privileged learners with access to personal mentors. With the expansion of mass education, such individualization became impractical, giving way to standardized curricula and uniform assessments. Subsequent advances in educational technologies, from computer-assisted instruction [19] to intelligent tutoring systems [20], sought to reintroduce personalization at scale, though often limited by rigid rule-based designs and high development costs.

While traditional large-scale education struggles to reconcile efficiency with personalization [21, 22], LLMs introduce a fundamentally new possibility [23, 24], capable of producing dynamic, context-aware, and individualized feedback that goes far beyond static rules or pre-programmed scripts. Unlike earlier systems, they can analyze student responses, diagnose potential misconceptions, and generate tailored guidance [25]. This capacity positions LLMs as powerful tools for scalable personalized learning, offering individualized support to learners across diverse educational environments. In this sense, LLMs may serve not merely as technological innovations, but as catalysts for reshaping the balance between standardization and personalization in education.

Consequently, evaluating LLMs for personalized learning requires more than measuring general model performance. It requires assessment frameworks capable of examining how models behave within realistic educational scenarios and how their pedagogical behaviors differ when supporting the same learner. This observation motivates the scenario-based evaluation framework proposed in this study.

# 3 Experimental Design

## 3.1 Quiz Data and Student Responses

To simulate a realistic post-class tutoring scenario, a set of quiz items was constructed based on an undergraduate data structures course. The objective was not to create a benchmark dataset for ranking model performance, but to establish a controlled educational setting in which differences in LLM tutoring behavior could be systematically examined. The quiz items covered a range of topics, including arrays, linked lists, stacks, queues, trees, graphs, and algorithmic complexity. They were deliberately kept concise to minimize ambiguity and facilitate detailed analysis of the complete tutoring process, including knowledge diagnosis, mastery inference, and personalized guidance generation. Although the examples used in this study are relatively simple, the proposed evaluation workflow is not restricted to this level of complexity and can be readily extended to more advanced educational materials and learning activities.

Each quiz item was paired with a student response and a correctness label. Given the quiz data, the LLMs were required to identify the underlying knowledge concepts, estimate the learner's level of mastery, and generate personalized recommendations for improvement. Notably, the models were not explicitly provided with the curriculum topics associated with each quiz item. Instead, they were required to extract and articulate the core concepts themselves [26]. This design more closely resembles authentic tutoring situations, where underlying concepts are rarely specified in advance and must be inferred from student work.

For transparency and reproducibility, all quiz items and responses were stored in JSON format, which explicitly encodes the question, student answer, and correctness label. This structured representation enables consistent input delivery across different LLMs, reduces ambiguity in data interpretation, and facilitates future replication and extension of the study. An illustrative JSON representation of the structured learning evidence is shown below:

```
{
  "id": 1,
  "question": "Explain the difference between a stack and a queue.",
  "student_answer": "A stack is FIFO and a queue is LIFO.",
  "correct": false
}
```

## 3.2 Prompt Design for Personalized Guidance

To facilitate systematic comparison of LLM behavior within the same educational scenario, a unified prompt was applied to all evaluated models. The prompt instructed each model to assume the role of an intelligent tutoring assistant specialized in data structures and to generate personalized learning guidance based on the dataset. By fixing the prompt content and structure, we ensured that all LLMs were evaluated under identical conditions, thereby minimizing the bias introduced by prompt engineering and allowing differences in diagnostic reasoning, mastery inference, and personalized guidance to be attributed primarily to the models themselves. Specifically, the prompt required the LLMs to perform three pedagogically oriented tasks:

- **Knowledge Diagnosis**: Identify the key knowledge concepts involved in each quiz item (e.g., recursion, linked list traversal, binary search tree insertion).
- **Mastery Inference**: Analyze student responses and classify each identified knowledge concept as mastered, partially understood, or not yet grasped.
- **Personalized Guidance Generation**: Provide individualized learning feedback, including (i) strengths and weaknesses associated with each concept, (ii) likely misconceptions or reasoning errors, (iii) specific, actionable learning suggestions, and (iv) recommended resources or practice activities.

A key feature of the prompt was its emphasis on structured and pedagogically meaningful output. Models were instructed to organize their responses using clear sections, subheadings, and bullet points, while adopting instructional language appropriate for university-level students. These requirements were incorporated as organization, clarity, interpretability, and actionability are important characteristics of effective tutoring support. The resulting outputs therefore are expected to resemble authentic tutoring feedback rather than generic question-answering responses. The complete prompt used in the experiments is provided below:

*You are an intelligent tutoring assistant specialized in data structures.*

*You will be given a set of data structures questions, each accompanied by a student's answer and whether the answer is correct or incorrect.*

*Your task is to:*

1. *Identify the key knowledge points involved in each question (e.g., recursion, linked list traversal, binary search tree insertion).*
2. *Analyze the student's performance across these questions to infer which knowledge areas are:*
   - *Mastered*
   - *Partially understood*
   - *Not yet grasped*
3. *Provide a detailed and personalized learning analysis, including:*
   - *Strengths and weaknesses for each identified knowledge point*
   - *Likely misconceptions or reasoning errors the student may have*
   - *Specific, actionable learning or review strategies tailored to the student's level*
   - *Recommended practice problems or resources, if applicable*

*Format your output into clear sections with appropriate subheadings and bullet points. Use precise, instructional language suitable for a university-level student.*

### 3.3 Prompt Design for Evaluation

Gemini was employed as an automated external evaluator, as LLM-based evaluators provide a practical mechanism for conducting large numbers of comparative assessments in a consistent and reproducible manner. For each comparison, Gemini received the same input data (quiz items, student answers, and correctness labels) together with two versions of personalized guidance generated by different LLMs. Its role was not to establish pedagogical ground truth, but to identify relative differences between alternative tutoring responses generated under the same learning scenario.

A unified evaluator prompt was designed and applied across all comparisons. The prompt instructed Gemini to act as a comparative evaluator focusing on relative quality across multiple pedagogically meaningful dimensions. Consistent with the motivation of this study, the objective was to characterize differences in tutoring behavior rather than to produce absolute judgments of educational quality. By emphasizing pairwise preference judgment rather than absolute scoring, the workflow reduced score calibration drift and improved discriminative consistency across different outputs. In contrast, absolute scoring often led to compressed rating distributions in which subtle but educationally important differences become difficult to distinguish.

Furthermore, Gemini was instructed to focus on semantic quality, logical coherence, pedagogical relevance, and instructional usefulness rather than superficial characteristics such as sentence length or formatting style. Each evaluation produced a categorical outcome: +1 (Model A preferred), −1 (Model B preferred), or 0 (tie). This design encourages balanced use of the decision space while enabling meaningful pedagogical differences to be reflected in the comparative outcomes. The complete evaluator prompt used in the experiments is provided below:

*You are a strict and consistent evaluator of personalized learning feedback.*

*## Task*

*You will be given:*

*1. A question set file containing:*
- *Data structures questions*
- *A student's answers*
- *Whether each answer is correct or incorrect*

*2. Model A's personalized guidance output for this student.*

*3. Model B's personalized guidance output for this student.*

*Your task:*

*1. Compare the output from Model A with the output from Model B.*

*2. Evaluate the contents relative to each other across the following five dimensions:*
- *Accuracy of knowledge diagnosis*
- *Identification of misconceptions*
- *Specificity and actionability of feedback*

- *Instructional clarity*
- *Appropriateness to the student's current level*

*3. Use relative judgment, not absolute scores.*

*4. Focus your comparison on semantic meaning, argument structure, logical coherence, evidence strength, and depth of insight, not on superficial details like sentence length, formatting, or stylistic phrasing.*

*5. After comparing all five dimensions, decide the overall winner:*
- *+1 → Model A is better overall for this pair.*
- *−1 → Model B is better overall for this pair.*
- *0 → They are equally good (or differences are negligible).*

*## Important Rules*

*1. Always use the full range of possible decisions.*

*2. If neither output is very good, still pick the better one unless they are truly equal.*

*3. Focus on differences, not similarities.*

*4. Be strict: small weaknesses should matter if the other output does better.*

*5. Do not output explanations or reasoning — only your decision.*

*## Output Format*

*Model A vs. Model B: +1/0/−1*

### 3.4 Methodological Considerations

Two design considerations are particularly important. First, the use of Gemini as an automated evaluator is intended to provide a scalable and consistent comparative signal rather than authoritative educational judgment. In educational assessment, even human evaluators may exhibit inconsistency, subjective interpretation, and variability across repeated evaluations. Within the proposed workflow, controlled prompting and fixed evaluation criteria are used to improve reproducibility and provide a stable basis for comparative analysis of LLM-generated tutoring behavior.

Second, the quiz correctness labels included in the dataset are not strictly required for the tutoring task itself. In principle, LLMs may also infer correctness as part of the diagnostic process before generating personalized guidance. Their inclusion in this study primarily serves to standardize experimental inputs and facilitate controlled comparison of diagnostic and tutoring behaviors across models. Future work may explore more open-ended settings in which correctness labels are withheld, thereby enabling simultaneous evaluation of grading, diagnosis, and tutoring capabilities.

More broadly, general-purpose benchmarks are primarily designed to measure model capability. Educational scenarios, however, additionally require understanding how models behave when supporting individual learners. Such behaviors may involve misconception identification, instructional clarity, actionability, appropriateness to learner level, and other pedagogical characteristics that are difficult to capture through conventional benchmark-oriented evaluation. The present study therefore serves as a proof-of-concept demonstration of how scenario-based evaluation can complement traditional benchmark-oriented assessment by providing a richer understanding of LLM behavior in personalized learning contexts.

## 4 Experiments

### 4.1 Workflow Overview

The workflow was designed to simulate an authentic post-class tutoring scenario while enabling systematic evaluation of multiple LLMs. Although this study employed a specific set of tutoring and evaluating models for demonstration, the workflow itself is model-agnostic, allowing future research to substitute alternative LLMs without requiring modification to the overall methodology.

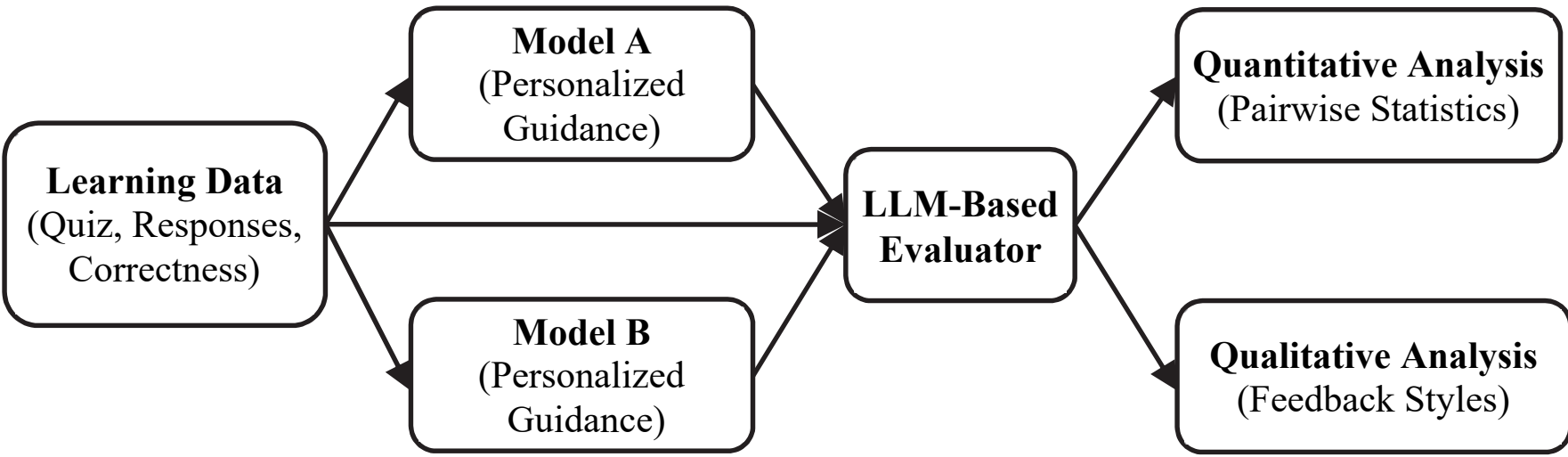

**Fig. 2.** Experimental workflow for examining LLM behavior in personalized learning.

As illustrated in Fig. 2, the experimental workflow consisted of four main stages: input preparation, personalized guidance generation, external evaluation, and analysis. In the first stage, a learning dataset was constructed, containing quiz items, student responses, and correctness labels. This input provided a standardized basis for subsequent evaluations, ensuring comparability across models. In the second stage, GPT-4o, DeepSeek-V3, and GLM-4.5 were each prompted with the same data and instructions. For each run, the three models produced personalized guidance that included knowledge concept extraction, mastery estimation, and targeted learning suggestions. To account for variation in generative outputs, multiple runs were conducted for each model. This design allows the evaluation to capture both model tendencies and generation variability within the same tutoring scenario.

In the third stage, the outputs were evaluated by Gemini for pairwise comparisons. For each comparison, Gemini received three inputs: (i) the learning data, (ii) Model A's output, and (iii) Model B's output. Based on this information, it produced relative judgments (+1, 0, −1) across five pedagogical criteria: accuracy of knowledge diagnosis, identification of misconceptions, specificity and actionability of feedback, instructional clarity, and appropriateness to the student's level. These criteria were selected as they reflect pedagogical characteristics that are central to personalized learning support.

Finally, the evaluation results were synthesized through two complementary forms of analysis. The quantitative analysis included pairwise win–tie–loss matrices, strength estimation with confidence intervals, and pairwise preference probabilities. The qualitative analysis examined representative case studies and stylistic differences in feedback. Together, these analyses revealed how different LLMs approached diagnosis, mastery inference, and personalized guidance under the same educational conditions.

### 4.2 Quantitative Analysis

In each run, all three models generated outputs based on the same learning data, after which three pairwise comparisons (GPT-4o vs. DeepSeek-V3, GPT-4o vs. GLM-4.5, and DeepSeek-V3 vs. GLM-4.5) were conducted. This procedure was repeated across 10 runs, resulting in a total of 30 pairwise evaluations. For each matchup, Gemini determined which model provided more favorable guidance, or whether the two outputs were of comparable quality. The results are summarized in Table 1 as win–tie–loss counts from the perspective of the row model.

**Table 1.** Pairwise win-tie-loss results among the three LLMs across 10 runs.

| Model | GPT-4o | DeepSeek-V3 | GLM-4.5 |
|---|---|---|---|
| **GPT-4o** | — | 5–2–3 | 9–1–0 |
| **DeepSeek-V3** | 3–2–5 | — | 7–2–1 |
| **GLM-4.5** | 0–1–9 | 1–2–7 | — |

Inspection of the win–tie–loss matrix suggests several relative tendencies under the proposed evaluation setting. GPT-4o was more frequently preferred over GLM-4.5 and also showed a moderate tendency to be preferred over DeepSeek-V3 in pairwise comparisons. DeepSeek-V3, in turn, was more frequently preferred over GLM-4.5 across most runs. These outcomes indicate that the proposed workflow can capture distinguishable pedagogical characteristics among different LLM outputs when they are applied to the same learning evidence.

While pairwise win–loss counts provide direct evidence of comparative preferences, they do not readily provide a unified characterization of relative model tendencies. To integrate results across all matchups, we employed the Bradley–Terry (BT) model [27], a probabilistic framework widely used for paired comparison data. In this model, each system is assigned a latent strength parameter $s$, and the probability of model $i$ being judged superior to model $j$ is defined as:

$$\mathrm{P}(i \succ j) = \frac{e^{s_i}}{e^{s_i} + e^{s_j}} \quad (1)$$

Based on the 30 pairwise comparisons, the estimated BT strengths are summarized in Table 2 and visualized in Fig. 3 using a forest plot [28]. The strength values were mean-centered for interpretability: positive values indicate above-average preference tendencies, while negative values indicate below-average preference tendencies. The estimated BT strengths provide a unified representation of the relative differences among the evaluated models. Notable, GPT-4o (BT = 0.95, 95% CI [0.191, 1.718]) exhibited the strongest comparative tendency, with its confidence interval lying entirely above zero, suggesting a statistically distinguishable preference tendency under the present tutoring scenario and evaluation setting.

**Table 2.** Bradley-Terry strength estimates with 95% confidence intervals for the three LLMs.

| Model | BT Strength | SE | 95% CI |
|---|---|---|---|
| GPT-4o | 0.954 | 0.390 | [0.191, 1.718] |
| DeepSeek-V3 | 0.364 | 0.356 | [−0.332, 1.061] |
| GLM-4.5 | −1.319 | 0.455 | [−2.210, −0.428] |

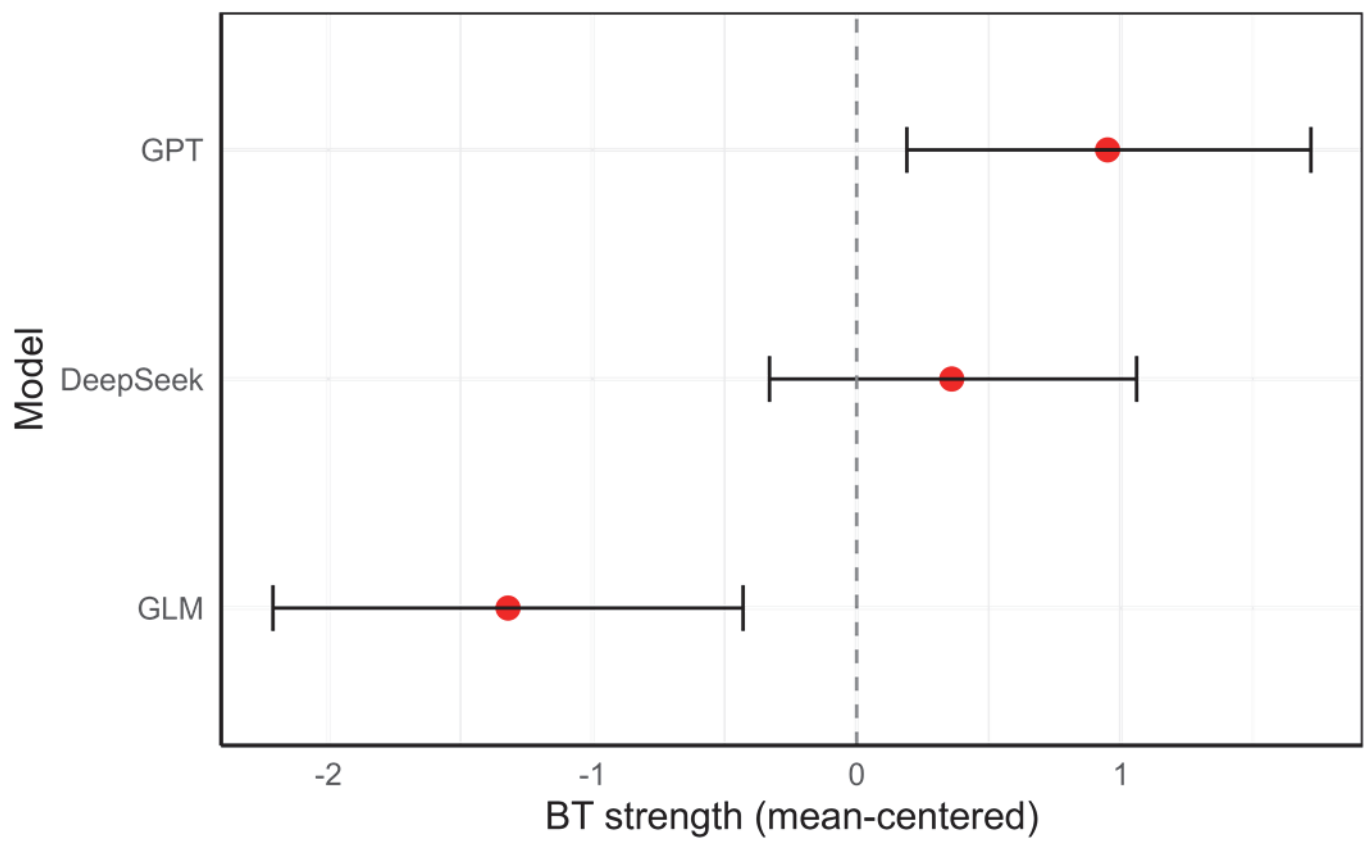

**Fig. 3.** Forest plot of Bradley-Terry strength estimates with 95% confidence intervals (CIs) for the evaluated LLMs. Red dots represent point estimates, and horizontal lines indicate 95% CIs.

DeepSeek-V3 (BT = 0.36, 95% CI [−0.33, 1.06]) occupied an intermediate position. Its confidence interval crossed zero, suggesting that although DeepSeek-V3 frequently produced competitive outputs, its relative advantage was not clearly distinguishable under the current experimental setting. GLM-4.5 (BT = −1.32, 95% CI [−2.21, −0.43]) showed comparatively less preferred outputs, with a negative strength estimate and a confidence interval fully below zero. Importantly, these values should be interpreted as comparative tendencies produced within the proposed workflow rather than universal rankings of model quality across broader educational contexts.

**Table 3.** Predicted pairwise preference probabilities estimated by the Bradley-Terry model.

| Model | GPT-4o | DeepSeek-V3 | GLM-4.5 |
|---|---|---|---|
| **GPT-4o** | — | 0.643 | 0.907 |
| **DeepSeek-V3** | 0.357 | — | 0.843 |
| **GLM-4.5** | 0.093 | 0.157 | — |

In addition to the win–loss counts, according to Eq. 1, the Bradley–Terry model can convert strength estimates into predicted pairwise preferences, offering a more interpretable summary of the comparative tendencies observed in the data. As shown in Table 3, GPT-4o was estimated to be preferred over DeepSeek-V3 with a probability of 64%, and over GLM-4.5 with a probability of 91%. Similarly, DeepSeek-V3, was estimated to be preferred over GLM-4.5 with a probability of 84%.

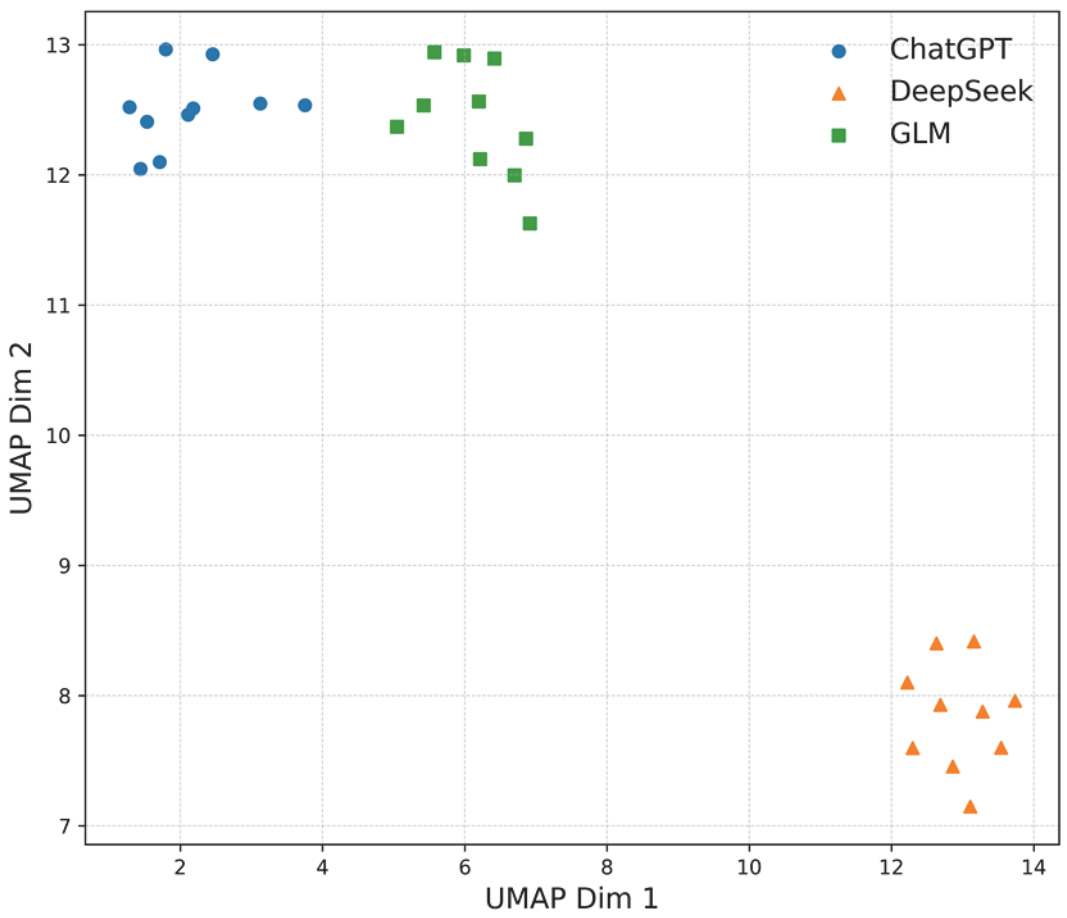

**Fig. 4.** 2D UMAP projection of sentence embeddings derived from 30 LLM-generated tutoring responses. Colors and marker shapes indicate different models.

Finally, the distributions of the model outputs were visualized using UMAP [29], based on sentence embeddings generated using the `all-MiniLM-L6-v2` encoder in the Sentence-Transformers framework [30]. In Fig. 4, GPT-4o and GLM-4.5 appeared relatively closer in the projected embedding space, whereas DeepSeek-V3 formed a comparatively separated cluster. Interestingly, this spatial pattern did not fully align with the comparative tendencies observed in the pedagogical evaluations. Models that produced semantically similar outputs were not necessarily evaluated similarly in terms of diagnosis quality, instructional usefulness, or actionability. This observation suggests that semantic similarity and pedagogical evaluation capture different aspects of LLM-generated tutoring responses and educational evaluation may require additional pedagogical dimensions that extend beyond textual similarity alone.

### 4.3 Qualitative Analysis

While quantitative comparisons provide useful summaries of relative tendencies, they may not fully capture nuanced differences in explanatory style and instructional strategy. Therefore, qualitative analysis was conducted to examine variations in feedback organization, diagnostic depth, instructional clarity, and actionability across models. The analysis aimed to illustrate how scenario-based educational evaluation can reveal distinct pedagogical behaviors in tutoring-oriented interactions.

**Clarity and Structure of Explanations.** GPT-4o frequently produced highly structured outputs with explicit headings, bullet points, and clearly separated sections for strengths, weaknesses, and recommendations. Such organization may enhance readability and help learners identify instructional points more efficiently. DeepSeek-V3 also provided systematic coverage of student performance, although some responses tended to provide more narrative explanations and the separation was sometimes less explicit. In contrast, GLM-4.5 tended to generate more narrative-style responses, which may appeal to learners who prefer more conversational forms of feedback.

**Depth of Knowledge Diagnosis.** GPT-4o often identified multiple related concepts within a single question and connected student errors to broader conceptual misunderstandings. DeepSeek-V3 also recognized many major misconceptions, although the level of diagnostic granularity appeared to vary across runs. In comparison, GLM-4.5 generally produced concise analyses that focused more heavily on answer correctness and core conceptual distinctions rather than detailed reasoning gaps.

**Specificity and Actionability of Feedback.** The models also differed in their preferred instructional strategies for supporting student learning. GPT-4o frequently provided concrete study suggestions, coding exercises, and references to external learning resources. While DeepSeek-V3 also generated actionable recommendations in many cases, some suggestions remained relatively general. In comparison, GLM-4.5 tended to emphasize broader review-oriented guidance, such as revisiting definitions or completing additional exercises, with less detailed instructional sequencing.

**Table 4.** Example of evidence-based learner diagnosis and personalized instructional guidance.

| Pedagogical Stage | Example Output |
|---|---|
| Knowledge Identification | DFS, BFS, Stack, Queue |
| Mastery Assessment | BFS mastered; DFS not yet grasped |
| Misconception Detection | DFS incorrectly associated with a queue |
| Recommendation | Review differences between DFS and BFS |
| Learning Resources | VisuAlgo, LeetCode exercises |

Beyond stylistic differences, an interesting observation was that the tutoring responses generated by different LLMs exhibited a broadly similar pedagogical reasoning structure. A representative example concerning the relationship among DFS, BFS, stacks, and queues is presented in Table 4. Despite variations in wording, organization, and level of detail, the models generally began by identifying the knowledge concepts involved in the questions, followed by an assessment of the learner's mastery status and the detection of potential misconceptions. Based on this diagnostic process, the models subsequently generated targeted learning recommendations and suggested relevant learning resources. Such behavior suggests the presence of an implicit learner-modeling process in which observable performance evidence is transformed into personalized instructional interventions.

### 4.4 Beyond Benchmarking

Traditional LLM evaluation is largely benchmark-oriented, emphasizing model ranking and direct performance comparisons. While such approaches are valuable for measuring general-purpose model competence, they provide limited insight into how models behave when performing complex educational tasks. From this perspective, the relatively focused tutoring scenario adopted in this study is a deliberate design choice rather than a methodological constraint, as the manageable set of post-class exercises enables detailed examination of the tutoring process. Expanding the evaluation into a large benchmark-style dataset would increase scale but could simultaneously obscure the pedagogical characteristics that the proposed framework seeks to expose.

The use of an external LLM evaluator serves a similar purpose. Rather than claiming pedagogical ground truth, it provides a consistent and scalable mechanism for identifying differences in LLM tutoring behavior. Within this framework, the resulting comparisons function as analytical signals that help characterize model behavior instead of definitive judgments of educational quality. Moreover, this study advocates a shift from benchmark-oriented evaluation toward scenario-based evaluation, which complements traditional benchmarks by examining how models behave in educational contexts.

## 5 Conclusion and Future Work

### 5.1 Major Findings

This study addresses a key issue in AI-enhanced education: how LLMs can be evaluated in pedagogically meaningful ways within authentic learning scenarios. In particular, we propose a scenario-based evaluation workflow for assessing LLMs in personalized learning support. Using a post-class tutoring scenario as an illustrative example, the framework examines how different LLMs analyze student responses, identify underlying concepts and misconceptions, infer mastery levels, and generate personalized guidance. Furthermore, to support systematic comparison, the workflow incorporates automated pairwise evaluation across multiple pedagogical dimensions.

The experimental results suggest that different LLMs exhibit distinguishable pedagogical behaviors under the proposed evaluation setting. More importantly, the study illustrates that educationally grounded evaluation workflows can reveal pedagogical differences that may not be fully captured by general benchmark-oriented evaluation alone. Overall, the primary contribution of this work lies not in establishing universal rankings of LLM capability, but in presenting a transparent, reproducible, and extensible methodology for examining LLM behavior in authentic educational scenarios.

### 5.2 Future Work

Future research may further extend and validate the proposed framework in several directions. First, more diverse educational data may be incorporated, including multiple disciplines, open-ended tasks, programming assignments, and multimodal learning materials. Second, future studies may explore additional pedagogical dimensions such as engagement, adaptivity, fairness, cognitive scaffolding, and long-term learning support. Third, cross-cultural studies may yield important insights into how LLMs can possibly adapt to different educational settings, including student backgrounds, learning habits, and instructional languages [31, 32]. Finally, human-in-the-loop validation [33] may be incorporated to collect feedback from students and instructors, providing valuable evidence regarding the perceived usefulness, pedagogical relevance, and educational impact of LLM-generated personalized guidance in real-world learning contexts.

**Acknowledgments.** The authors acknowledge the use of AI-assisted writing tools during the preparation of this manuscript. The authors retain full responsibility for the research design, experimental workflow, interpretations, and final content of the paper.

## References

1. GPT-4o, https://openai.com/index/hello-gpt-4o, last accessed 2025/09/01
2. DeepSeek-AI Team: DeepSeek-V3 technical report. arXiv:2412.19437v2 (2025)
3. GLM-4.5 Team: GLM-4.5: Agentic, reasoning, and coding (ARC) foundation models. arXiv:2508.06471 (2025)
4. Brown, T., et al.: Language models are few-shot learners. In: 34th International Conference on Neural Information Processing Systems, pp. 1877–1901 (2020)
5. Fütterer, T., et al.: ChatGPT in education: Global reactions to AI innovations. Scientific Reports **13**, Article No. 15310 (2023)
6. Shemshack, A., Spector, J.: A systematic literature review of personalized learning terms. Smart Learning Environments **7**, Article No. 33 (2020)
7. Zhang, L., Basham, J., Yang, S.: Understanding the implementation of personalized learning: A research synthesis. Educational Research Review **31**, Article No. 100339 (2020)
8. Kasneci, E., et al.: ChatGPT for good? On opportunities and challenges of large language models for education. Learning and Individual Differences **103**, Article No. 102274 (2023)
9. Liang, P., et al.: Holistic evaluation of language models. Transactions on Machine Learning Research, https://openreview.net/forum?id=iO4LZibEqW (2023)
10. Adeshola, I., Adepoju, A.: The opportunities and challenges of ChatGPT in education. Interactive Learning Environments **32**(10), 6159–6172 (2024)
11. Chen, A., Wei, Y., Le, H., Zhang, Y.: Learning by teaching with ChatGPT: The effect of teachable ChatGPT agent on programming education. British Journal of Educational Technology **57**(1), 163–184 (2026)
12. Liao, J., Zhong, L., Zhe, L., Xu, H., Liu, M., Xie, T.: Scaffolding computational thinking with ChatGPT. IEEE Transactions on Learning Technologies **17**, 1628–1642 (2024)
13. Ouyang, F., Guo, M., Zhang, N., Bai, X., Jiao, P.: Comparing the effects of instructor manual feedback and ChatGPT intelligent feedback on collaborative programming in China's higher education. IEEE Transactions on Learning Technologies **17**, 2173–2185 (2024)
14. Bloxham, S., Boyd, P.: Developing effective assessment in higher education: A practical guide. Open University Press, Maidenhead (2007)
15. Gemini 2.5, https://deepmind.google/models/gemini, last accessed 2025/09/01
16. Zheng, L., et al.: Judging LLM-as-a-judge with MT-Bench and Chatbot Arena. In: 37th International Conference on Neural Information Processing Systems, pp. 46595–46623 (2023)
17. Gu, J., et al.: A survey on LLM-as-a-judge. The Innovation **7**(6), Article No. 101253 (2026)

18. Grant, P., Basye, D.: Personalized learning: A guide for engaging students with technology. International Society for Technology in Education, Eugene (2014)
19. Chambers, J., Sprecher, J.: Computer assisted instruction: Current trends and critical issues. Communications of the ACM **23**(6), 332–342 (1980)
20. Anderson, J., Boyle, C., Reiser, B.: Intelligent tutoring systems. Science **228**(4698), 456–462 (1985)
21. Black, P., Wiliam, D.: Assessment and classroom learning. Assessment in Education: Principles, Policy & Practice **5**(1), 7–74 (1998)
22. Hattie, J., Timperley, H.: The power of feedback. Review of Educational Research **77**(1), 81–112 (2007)
23. Sharma, S., Mittal, P., Kumar, M., Bhardwaj, V.: The role of large language models in personalized learning: A systematic review of educational impact. Discover Sustainability **6**, Article No. 243 (2025)
24. Pataranutaporn, P., et al.: AI-generated characters for supporting personalized learning and well-being. Nature Machine Intelligence **3**, 1013–1022 (2021)
25. Yuan, B., Hu, J.: Generative AI as a tool for enhancing reflective learning in students. In: 2025 IEEE International Conference on Teaching, Assessment, and Learning for Engineering. IEEE, Macao (2025)
26. de la Torre, J.: DINA model and parameter estimation: A didactic. Journal of Educational and Behavioral Statistics 34(1), 115–130 (2009)
27. Bradley, R., Terry, M.: Rank analysis of incomplete block designs: I. The method of paired comparisons. Biometrika **39**(3/4), 324–345 (1952)
28. Lewis, S., Clarke, M.: Forest plots: Trying to see the wood and the trees. British Medical Journal **322**(7300), 1479–1480 (2001)
29. Mclnnes, L., Healy, J., Melville, J.: UMAP: Uniform manifold approximation and projection for dimension reduction. arXiv:1802.03426 (2018)
30. Reimers, N., Gurevych, I.: Sentence-BERT: Sentence embeddings using Siamese BERT-Networks. In: 2019 Conference on Empirical Methods in Natural Language Processing, pp. 3982–3992. ACL, Hong Kong (2019)
31. Shan, X., Xu, Y., Wang, Y., Lin, Y., Bao, Y.: Cross-cultural implications of large language models: An extended comparative analysis. In: 26th International Conference on Human-Computer Interaction, LNCS 15375, pp. 106–118. Springer, Washington DC (2024)
32. Hawk, T., Shah, A.: Using learning style instruments to enhance student learning. Decision Sciences Journal of Innovative Education **5**(1), 1–19 (2007)
33. Armfield, D., et al.: Avalon: A human-in-the-loop LLM grading system with instructor calibration and student self-assessment. In: 26th International Conference on Artificial Intelligence in Education, CCIS 2592, pp. 111–118. Springer, Palermo (2025)